\documentclass[lettersize,journal]{IEEEtran}
\usepackage{amsmath,amsfonts}
\usepackage{algorithmic}
\usepackage{algorithm}
\usepackage{array}
\usepackage[caption=false,font=normalsize,labelfont=sf,textfont=sf]{subfig}
\usepackage{textcomp}
\usepackage{stfloats}
\usepackage{url}
\usepackage{verbatim}
\usepackage{graphicx}
\usepackage{cite}
\usepackage{stackengine}
\stackMath
\usepackage[colorlinks=true, linkcolor=blue, citecolor=blue]{hyperref}
\DeclareGraphicsExtensions{.png, .pdf}
\DeclareGraphicsExtensions{.pdf, .png}

\begin{document}

\title{SPINE Gripper: A Twisted Underactuated Mechanism-based Passive Mode-Transition Gripper}

\author{JaeHyung Jang, JunHyeong Park, Joong-Ku Lee, and Jee-Hwan Ryu 
        % <-this % stops a space
\thanks{This work was supported in part by the Ministry of Trade, Industry and Energy (MOTIE), Republic of Korea, under Grant No. 00416440, and in part by the National Research Foundation of Korea (NRF) funded by the Korea government (MSIT) under Grant No. RS-2025-02213804. \textit{(JaeHyung Jang and JunHyeong Park contributed equally to this work.) (Corresponding
author: Jee-Hwan Ryu.)}}

\thanks{JaeHyung Jang, Joon-ku Lee, and Jee-Hwan Ryu are with the Department of Civil and Environmental Engineering, Korea Advanced Institute of Science and Technology, Daejeon 34141, South Korea. (e-mail: jhjang.kd@kaist.ac.kr; iamjoong9@kaist.ac.kr; jhryu@kaist.ac.kr).}

\thanks{JunHyeong Park is with the Robotics Program, Korea Advanced Institute of Science and Technology, Daejeon 34141, South Korea (e-mail: jamesjun2000@kaist.ac.kr).}}
% <-this % stops a space

% The paper headers
%% \markboth{Journal of \LaTeX\ Class Files,~Vol.~14, No.~8, August~2021}%

% {Shell \MakeLowercase{\textit{et al.}}: A Sample Article Using IEEEtran.cls for IEEE Journals}

%\IEEEpubid{0000--0000/00\$00.00~\copyright~2021 IEEE}

% Remember, if you use this you must call \IEEEpubidadjcol in the second
% column for its text to clear the IEEEpubid mark.

\maketitle

\begin{abstract}
This paper presents a single-actuator passive gripper that achieves both stable grasping and continuous bidirectional in-hand rotation through mechanically encoded power transmission logic. Unlike conventional multifunctional grippers requiring multiple actuators, sensors, or control-based switching, the proposed gripper transitions between grasping and rotation solely according to applied input torque magnitude. The key enabler of this behavior is a Twisted Underactuated Mechanism (TUM), which generates non-coplanar motions, axial contraction and rotation, from a single rotational input while producing identical contraction regardless of rotation direction. A friction generator mechanically defines torque thresholds that govern passive mode switching, enabling stable grasp establishment before autonomously transitioning to in-hand rotation without sensing or active control. Analytical models describing the kinematics, elastic force generation, and torque transmission of the TUM are derived and experimentally validated. The fabricated gripper evaluated through quantitative experiments on grasp success, friction-based grasp force regulation, and bidirectional rotation performance. System-level demonstrations, including bolt manipulation, object reorientation, and manipulator-integrated tasks driven solely by wrist torque, confirm reliable grasp–rotate transitions in both rotational directions. These results demonstrate that non-coplanar multifunctional manipulation is realized through mechanical design alone, establishing mechanically encoded power transmission logic as a robust alternative to actuator- and control-intensive gripper architectures.

\end{abstract}


\section{Introduction}

% (c)에 대한 설명도 추가
\IEEEPARstart{R}{obotic} grippers are fundamental elements of robotic manipulation systems, allowing robots to physically interact with objects of varying shapes, sizes, and materials. Consequently, grippers play a critical role across a wide range of applications, including industrial automation, warehouse logistics, service robotics, and agricultural harvesting \cite{belanche2020service, dupont2021decade, elfferich2024berrytwist, barakazi2022use}. In many of these applications, grippers are required to perform more than simple opening and closing motions. Practical manipulation tasks often demand stable handling of objects with diverse properties, adjustment of object orientation according to task requirements, and fine, contact-rich manipulation after grasping. Such requirements call for advanced capabilities such as adaptive grasping and in-hand manipulation, while maintaining high reliability in complex and unstructured environments.

Conventional approaches to realizing these multifunctional behaviors typically rely on multiple actuators, distributed sensors, and sophisticated control architectures. While effective, these solutions inevitably increase system volume, weight, and power consumption, and introduce additional mechanical and electrical failure modes, reducing overall robustness and reliability \cite{kim2024inherently,tejada2025review, hegde2023sensing}. As a result, there is a growing demand for gripper design strategies that can achieve multiple manipulation functions while preserving structural simplicity and minimizing actuation and control complexity.

\begin{figure}[!t]  
\centering
\includegraphics[width=8cm]{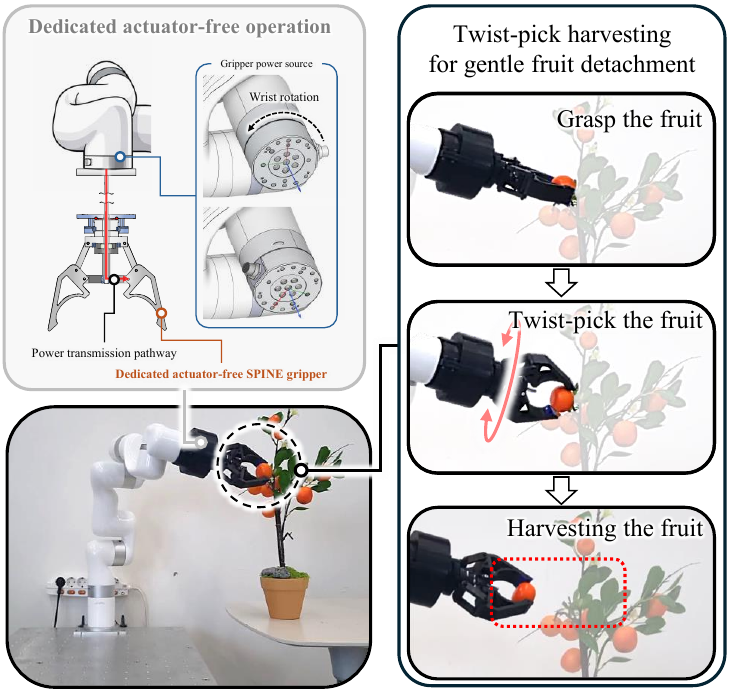}
\caption{Single actuator-based PassIve Non-coplanar degree of freedom modE transition (SPINE) gripper: Passive mode-transition gripper with mechanically encoded power transmission logic for grasp and in-hand rotation in fruit harvesting, exploiting the manipulator wrist's rotational motion without dedicated actuators or control electronics.}
\label{fig:fig17}
\end{figure}

%Single actuator based multi-functional mechanisms have emerged as a promising approach to address these challenges by embedding multiple behaviors within the physical structure itself. This approach simplifies the system structure and control architecture, thereby enhancing the overall reliability of robotic platforms. 

To meet this demand for mechanical simplicity and high reliability, several studies have proposed underactuated mechanisms based multi-functional grippers. Ko et al. presented a tendon-driven robot gripper that passively switches crawler surfaces with a single actuator to grasp thin objects\cite{ko2020tendon}. Yoon et al. developed underactuated fingers that accommodate various object shapes by combining variable-length links with tendon structures\cite{yoon2022elongatable}. Additionally, Hussein et al. proposed an underactuated gripper that mechanically implements desired equivalent stiffness and contact distribution through compliant joint properties and modular architecture\cite{hussain2018modeling}. Baril et al. introduced prosthetic grippers that passively select operation modes or regulate force distribution using mechanical levers and selectors\cite{baril2013design}. Meanwhile, Seino et al. proposed a fully passive gripper (PALGRIP) containing no active actuators, demonstrating that grasping and releasing occur solely through structural constraints and contact surface friction\cite{seino2025passive}. Furthermore, research on adaptive underactuated grippers has been reported, embedding object shape adaptability within the structure itself using variable-length links and elastic elements\cite{borisov2022reconfigurable, kragten2011stable}.

\begin{figure*}[!t]
\centering
\includegraphics[width=17cm]{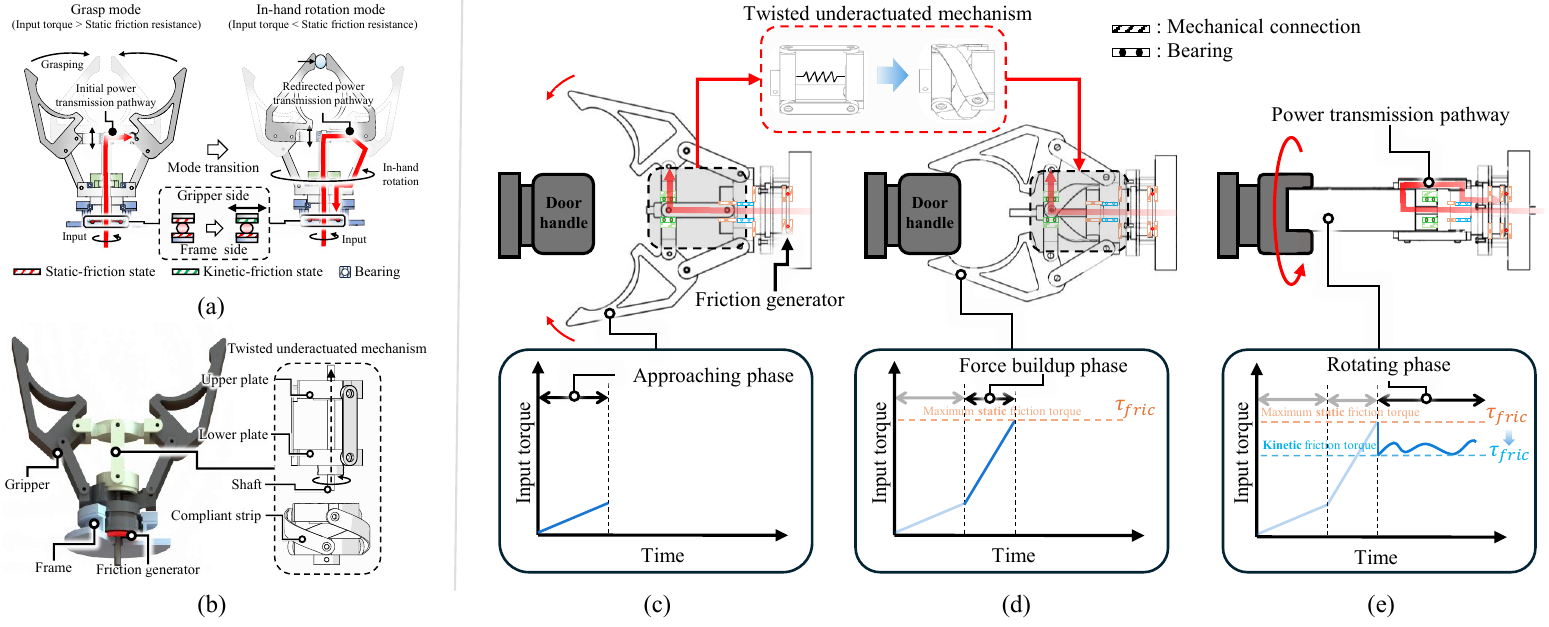}
\caption{Functional components and mechanically encoded power transmission behavior of the SPINE gripper during door handle manipulation. (a) The SPINE gripper exhibits friction-governed transition between grasping and in-hand rotation, with mode switching determined by torque thresholds. (b) Structural elements include the twisted underactuated mechanism (TUM), gripper, friction generator, and frame. (c) Approaching phase: the gripper moves toward the door handle without generating grasping force. (d) Force buildup phase: contact induces grasping force as input torque increases. (e) Rotating phase: input torque exceeding maximum static friction initiates rotation while maintaining stable grasp; grasping force remains proportional to kinetic friction torque as the system transitions to continuous rotation.}
\label{fig:fig2}
\end{figure*}

While above-mentioned studies demonstrated that underactuated and passive mechanisms effectively improve adaptability and reduce actuation and control complexity, their use has largely been confined to enhancing grasp conformity, force distribution, or contact stability. In most cases, underactuation is exploited to passively adapt finger motion to object geometry, rather than to enable multiple, functionally distinct manipulation modes within a single mechanism. As a result, extending such approaches to support fundamentally different manipulation functions, particularly the combination of stable grasping and continuous in-hand rotation in practical manipulation tasks\cite{elfferich2022soft, prattichizzo2016grasping, rapacki2009underactuated}, remains a significant challenge.

Despite their importance, achieving both functions with a single actuator is fundamentally difficult because grasping and in-hand rotation involve motion along non-coplanar directions. Nishimura et al. demonstrated that grasping and in-hand rotation can be passively coordinated using a single actuator by exploiting force-magnitude–based mode switching\cite{nishimura20221}. However, their mechanism inherently supports rotation in only one direction due to sign-coupled force transmission, which restricts its applicability in tasks requiring directional reversals.

To overcome these limitations, we present a Single-actuator PassIve Non-coplanar degree-of-freedom modE-transition (SPINE) gripper that enables stable grasping and continuous bidirectional in-hand rotation through a mechanically encoded power transmission logic, direction-invariant contraction mechanism (Fig. \ref{fig:fig17}). Unlike conventional active-type multi-functional mechanisms, the proposed gripper achieves this functionality through a mechanically encoded power transmission logic rather than through sensing or active control. At the core of the system is a Twisted Underactuated Mechanism (TUM) that generates two non-coplanar motions, axial contraction and rotation, from a single rotational input, and produces identical contraction regardless of the input rotation direction. This property enables a passive mode-switching mechanism in which the transition between grasping and in-hand rotation is governed solely by the magnitude of the input torque. Through analytical modeling, mechanical characterization, and prototype demonstrations, we show that the proposed gripper achieves reliable multifunctional manipulation with a minimal hardware configuration. Furthermore, when mounted on a robotic manipulator and driven exclusively by wrist torque, the system functions as an actuator-free end effector capable of stable grasping and continuous bidirectional rotation, while maintaining structural simplicity and robustness.

% The core of the system is the Twisted Underactuated Mechanism, which provides two non coplanar motions, contraction and rotation, and generates contraction regardless of the direction of the applied rotation. This property enables a mechanically encoded switching mechanism that transitions between grasp and rotation purely through the magnitude of the input torque, without relying on sensors or active control. Through analytical modeling, mechanical characterization, and prototype demonstrations, we verify that the SPINE gripper achieves reliable multi-functional manipulation with a minimal hardware configuration. Furthermore, when mounted on a robotic manipulator and driven solely by the wrist actuator, the system functions as an actuator free gripper capable of both grasping and bidirectional rotation, enabling versatile object handling while maintaining structural simplicity.

The structure of this paper is organized as follows. Section \ref{sec:section2} introduces the overall system architecture, operating principles, and passive transition scenarios between grasping mode and in-hand rotation mode. Section \ref{sec:section3} presents the theoretical principles of the TUM and the kinematic model of the compliant strip structure. Section \ref{sec:section4} provides experimental validation of the theoretical predictions. Section \ref{sec:section5} describes in detail the design and fabrication process of the SPINE gripper prototype. Section \ref{sec:section6} validates the quantitative feasibility of the proposed functionalities through experiments including grasp success conditions, friction-grasping force relationships, and rotation performance evaluation. Section \ref{sec:section7} demonstrates various task scenarios utilizing a system integrated with a robotic manipulator, confirming performance and versatility in real-world environments. Section \ref{sec:section8} summarizes the research findings and suggests future research directions.

\section{Concept of the SPINE gripper}
\label{sec:section2}
This section presents the core concept of the SPINE gripper, which achieves both grasping and in-hand rotation through mechanically encoded power transmission logic driven by a single actuator. Instead of employing additional actuators, sensors, or control algorithms, the gripper alters its power transmission pathway according to applied input torque (Fig. \ref{fig:fig2}(a)), with predefined torque thresholds and mechanical constraints determining whether input motion routes toward grasping or in-hand rotation. 

As a result, this approach enables multifunctional manipulation through structural design alone. Unlike conventional grippers that employ multiple actuators, sensors, or control algorithms for multifunctional manipulation, the SPINE gripper generates non-coplanar degrees of freedom through passive component interactions. By embedding decision logic directly into the mechanical architecture, this approach achieves stable grasping and continuous in-hand rotation with minimal hardware, demonstrating that structural design alone can support reliable multi-functional operation.

% This section presents the core concept of the SPINE gripper, which achieves both grasping and in-hand rotation through mechanically encoded power transmission logic driven by a single actuator. Instead of employing additional actuators, sensors, or control algorithms, the SPINE gripper alters its power transmission pathway according to the applied input torque (Fig. \ref{fig:fig2}(a)). Predefined torque thresholds and mechanical constraints embedded in the structure determine whether the input motion is routed toward grasping or in-hand rotation.

% As a result, this approach enables multifunctional manipulation through structural design alone. Conventional grippers typically realize multiple functions by combining several actuators or relying on control-based mode switching, increasing system complexity, wiring, and failure risk. In contrast, the SPINE gripper generates non-coplanar degrees of freedom through passive interactions among its components, allowing motion modes to transition naturally as torque increases. By embedding decision logic directly into the mechanical architecture, the gripper achieves stable grasping and continuous in-hand rotation with minimal hardware, demonstrating that mechanical design alone supports reliable multi-functional operation.

\begin{figure*}[!t]
\centering
\includegraphics[width=16.5cm]{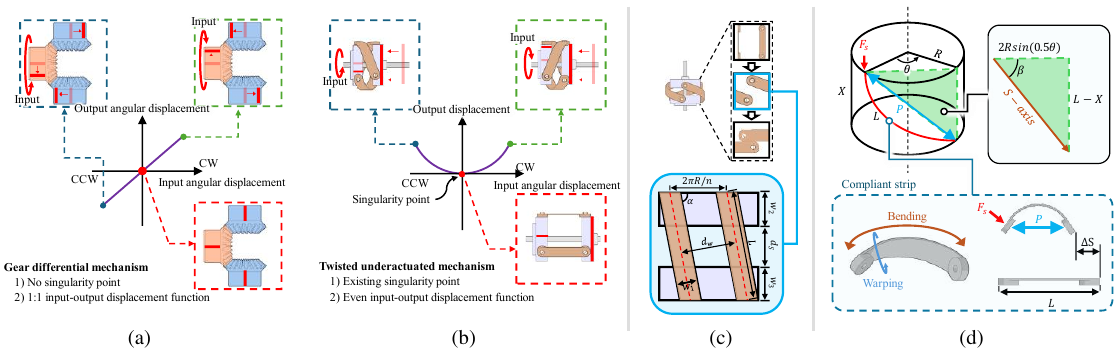}
\caption{Bidirectional behavior, stroke geometry, and deformation characteristics of the twisted underactuated mechanism (TUM). 
(a) A conventional gear differential exhibits no singularity, maintaining a fixed input-output displacement relationship.
(b) The TUM contains a singularity enabling unidirectional contraction under bidirectional input rotation.
(c) Strip spacing variation under twisting modeled using two-dimensional geometry. 
(d) Geometric and deformation characteristics relating rotation $\theta$, contraction $X$, and force projection along the $s$ axis, with bending-dominated deformation governing elastic force $F_s$.}
\label{fig:fig30}
\end{figure*}

\subsection{Functional Components of the SPINE Gripper} \label{sec:section2-1}

Passive mode switching in the SPINE gripper arises from mechanically encoded power transmission logic rather than active control. This mechanism comprises four primary components: a twisted underactuated mechanism (TUM), gripper, friction generator, and frame (Fig. \ref{fig:fig2}(b)). These components form an interdependent mechanical network where torque is transmitted, redirected, or released based on structural interactions, implementing the system's switching logic.

The TUM is the core element responsible for generating non-coplanar motion. Based on a screw-like mechanism, it converts rotational input into linear contraction. By removing the rotational constraint on the contracting element, it simultaneously enables contraction and rotation, forming a 2-DOF underactuated mechanism. This dual motion capability allows the same input torque to drive either grasping or in-hand rotation depending on downstream constraints. Detailed analysis is provided in Section \ref{sec:section3-1}.

The gripper employs a modified slider-crank mechanism that transforms TUM contraction into finger closing motion. Its rigid-body structure transmits force efficiently while allowing free rotation once the system transitions to rotation mode.

The friction generator defines the torque threshold for mode switching by frictionally coupling the gripper to the frame. Below the static friction limit, torque drives TUM contraction and grasping. When torque exceeds this limit, friction slips, disengaging the coupling and redirecting torque to rotate the gripper. Mode switching is therefore governed passively by torque magnitude alone.

The frame anchors the system and enforces mode-specific mechanical constraints. During grasping, it prevents gripper rotation while allowing TUM contraction. During in-hand rotation, it constrains all motions except gripper rotation, ensuring stable bidirectional rotation without active control.

\subsection{Operational phases of the SPINE Gripper} \label{sec:section2-2}

Because the SPINE gripper relies entirely on mechanically encoded power transmission logic, its behavior is explained through operational phases that describe the redistribution of input torque through the structure. These phases clarify the passive switching logic and the roles of elastic deformation, frictional thresholds, and mechanical constraints.
The SPINE gripper operates through three sequential phases: approaching, force buildup, and rotating. Each phase corresponds to a distinct mechanical state defined by interactions among the TUM, gripper, and friction generator (Fig. \ref{fig:fig2}).
In the approaching phase (Fig. \ref{fig:fig2}(c)), input torque causes the TUM to contract through elastic twisting, actuating the slider-crank mechanism and closing the gripper toward the object. During this motion, elastic deformation gradually increases the input torque, enabling a smooth approach without abrupt force application.
The force buildup phase begins upon fingertip contact (Fig. \ref{fig:fig2}(d)). Increased frictional resistance causes input torque to rise rapidly, generating a stable grasp. Torque continues increasing until reaching the static friction threshold set by the friction generator, passively limiting grasping force without sensors or feedback control.
In the rotating phase (Fig. \ref{fig:fig2}(e)), applied torque exceeds the friction threshold, transitioning the coupling between the gripper and frame from static to kinetic friction. Torque is then redirected to rotate the gripper while maintaining the established grasp force, with kinetic friction characteristics ensuring consistent rotation despite external torque fluctuations.
These phases demonstrate that the SPINE gripper achieves reliable transitions between grasping and in-hand rotation through purely mechanical interactions, validating mechanically encoded power transmission logic as a robust alternative to control-dependent multifunctional grippers.

\section{Design of the Twisted Underactuated Mechanism} \label{sec:section3}

Generating two-degree-of-freedom (DOF) non-coplanar motions from a single rotational input poses a fundamental challenge in underactuated mechanism design. Conventional power transmission elements such as differential gears and ball screws can theoretically achieve such configurations, but their structural characteristics, optimized for continuous power transmission, inherently produce odd-function input-output relationships that limit passive mode-switching to unidirectional operation. To overcome this constraint, we introduce the twisted underactuated mechanism (TUM), which simultaneously generates axial and rotational outputs from a single rotary input. Although demonstrated here through the SPINE gripper, the TUM serves as a mechanism-level enabler that allows the proposed gripper to achieve non-coplanar motion under a single passive transmission pathway. This section presents the theoretical foundation of the TUM, deriving analytical models for its kinematic and force transmission properties and providing design guidelines for broader implementation in underactuated mechanisms.

\subsection{Bidirectional capability of Twisted Underactuated Mechanism} \label{sec:section3-1}

Bidirectional in-hand rotation is fundamental for manipulation tasks such as tightening, loosening, and object reorientation. Conventional passive grippers based on differential mechanisms exhibit symmetric input–output rotation (Fig. \ref{fig:fig30}(a)): clockwise input produces clockwise output, and vice versa. This odd-function behavior transmits rotation only through aligned motion, resulting in effectively unidirectional manipulation once grasping occurs.

The TUM overcomes this limitation through a twisted compliant architecture. The mechanism consists of a bottom plate fixed to the input shaft, a top plate supported by a linear bushing allowing rotation and translation, and compliant strips connecting the two plates. Rotation of the bottom plate twists the strips, inducing linear contraction of the top plate. Crucially, this contraction occurs identically under both clockwise and counterclockwise rotation. As shown in Fig. \ref{fig:fig30}(b), the resulting input–output displacement is an even function with singularities, indicating directionally symmetric contraction.
This behavior enables identical grasping under opposite torque directions while permitting free bidirectional in-hand rotation after grasping. Unlike gear-based passive mechanisms fundamentally constrained to directional coupling, the TUM decouples contraction from rotation direction, substantially expanding the manipulation repertoire of passive grippers and establishes twisted compliant structures as a powerful alternative to conventional power transmission based underactuated mechanisms.

\subsection{Stroke design of the Twisted Underactuated Mechanism} \label{sec:section3-2}

The compliant strips provide the kinematic linkage that enables contraction through twisting deformation. Their geometric design therefore directly determines the operational stroke of the TUM, and these stroke constraints directly determine the achievable grasping range of the SPINE gripper.

For structural balance, the strips are arranged symmetrically and uniformly around the mounting plate. During twisting, the distance between adjacent strips decreases and eventually leads to interference, which defines the maximum operational range. As shown in Fig. \ref{fig:fig30}(c), the angular inclination $\alpha$ of each strip relative to the plate is given by
\begin{align}
\alpha = \arcsin{\frac{w_2+w_3+d_s}{L}}
\label{eqn:eqn1}
\end{align}
where $w_2$ and $w_3$ denote the thicknesses of the top and bottom plates, respectively, $d_s$ is the inter-plate distance, and $L$ is the strip length.

With $N$ strips placed on a plate of radius $R$, the distance between adjacent strip joints is
\begin{align}
d_w = \frac{2\pi R}{N}\sin{\alpha}
= \frac{2\pi R(w_2+w_3+d_s)}{NL}.
\label{eqn:eqn2}
\end{align}
This relation indicates that increasing the number of strips reduces the allowable stroke. In practice, adjacent strips contact each other through curved surfaces rather than parallel centerlines, which introduces instability in torque transmission. To avoid this, the TUM is designed such that strip interference does not occur even when the top and bottom plates fully contact ($d_s=0$). The resulting design constraint is
\begin{align}
     \frac{2\pi R(w_2+w_3+d_s)}{NL}=\frac{2\pi R(w_2+w_3)}{NL}\geq w_1
    \label{eqn:eqn3}
\end{align}
where $w_1$ is the strip width. By rearranging the remaining terms, we derive the design condition for the TUM radius as a function of the compliant strip specifications defined as follows:
\begin{align}
R \geq \frac{w_1NL}{2\pi (w_2+w_3)}.
\label{eqn:eqn3-1}
\end{align}

\subsection{Elastic Stiffness design of the Twisted Underactuated Mechanism} \label{sec:section3-3}

\begin{figure}[!t]
\centering
\includegraphics[width=7cm]{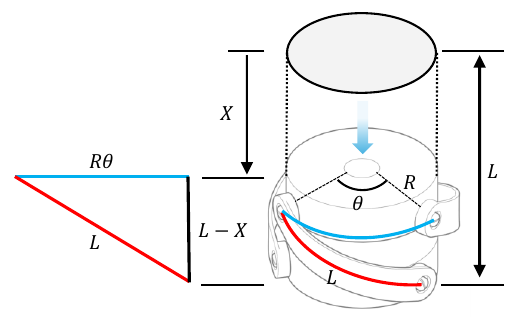}
\caption{Schematic representation for kinematic modeling of the TUM.}
\label{fig:fig6}
\end{figure}

The compliant strips are fabricated from elastic materials with high slenderness to generate restorative forces under twisting. These elastic forces determine both the contraction force of the TUM and the minimum friction threshold required for stable grasping before mode transition in the SPINE gripper.

As shown in Fig. \ref{fig:fig30}(d), twisting deformation induces both bending and warping; however, bending dominates the elastic response. Accordingly, the elastic energy is modeled solely based on bending. The axial elastic force of the TUM, $F_s$, is obtained by projecting the strip’s bending-induced elastic force $P$ onto the axial direction:
\begin{align}
F_s = P\cos{\beta}
\label{eqn:eqn4}
\end{align}
where $\beta$ is the angle between the strip and the axial direction.

The bending behavior of a compliant strip is governed by
\begin{align}
EI\frac{d^2y}{ds^2} + Py(s) = 0
\label{eqn:eqn5}
\end{align}
where $E$ is Young’s modulus, $I$ is the second moment of area, and $y(s)$ is the transverse deflection. Solving this equation yields
\begin{align}
    y(s) = A\sin({\sqrt\frac{P}{EI}\cdot s})=A\sin({\frac{\pi}{L-\Delta S}\cdot s}),
    \label{eqn:eqn6}
\end{align}
where $A$ is the maximum height of the strip. Through this, the height of the link is expressed in the form of a trigonometric function, and it is represented at Fig. \ref{fig:fig30}(d) and (\ref{eqn:eqn6}) that the distance between the endpoints of the strip corresponds to half a period. This relationship is expressed in the following equation and summarized for $P$:
\begin{align}
    \sqrt\frac{P}{EI} = \frac{\pi}{L-\Delta S}
    \label{eqn:eqn7}
\end{align}

\begin{align}
    P = (\frac{\pi}{L-\Delta S})^2\cdot EI.
    \label{eqn:eqn8}
\end{align}
From the triangle geometry in Fig. \ref{fig:fig30}(d), the relationship between contraction, rotation, and orientation angle is given by
\begin{align}
(L-\Delta S)^2 &= L^2-(R\theta)^2+4R^2\sin^2\left(\frac{\theta}{2}\right)
\label{eqn:eqn9}
\end{align}
\begin{align}
\beta &= \arctan{\frac{2R\sin(0.5\theta)}{L-X}}.
\label{eqn:eqn10}
\end{align}
Combining these relations yields the axial elastic force of a single strip as a function of $\theta$:
\begin{align}
\begin{split}
    F_s = \frac{EI \pi^2}{L^2 - (R \theta)^2 + 4R^2 \sin^2 \left( \frac{\theta}{2} \right)}
    \\\cdot \frac{R^2 \sin(\theta)}{\sqrt{2R^2} - 2R^2 \cos(\theta) + L^2 - (R \theta)^2}.
    \label{eqn:eqn11}
    \end{split}
    \end{align}   
This expression quantitatively captures the dependence of elastic force on material properties, geometry, and rotation.

\begin{figure}[!t]
\centering
\includegraphics[width=7cm]{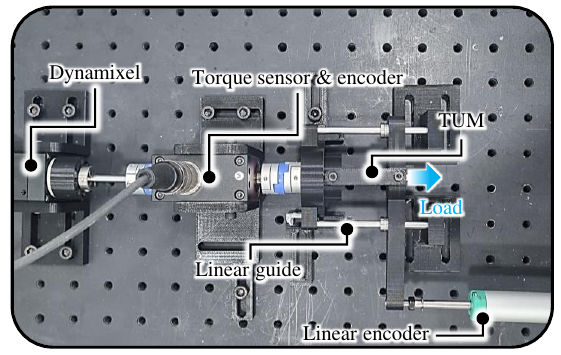}
\caption{Experimental setup to evaluate kinematics, elastic force and Jacobian-based motion anlaysis of the TUM.}
\label{fig:fig7}
\end{figure}

\subsection{Kinematic Modeling of the Twisted Underactuated Mechanism} \label{sec:section3-4}

For kinematic analysis, the TUM is idealized by modeling the plates as zero-thickness disks and the compliant strips as zero-thickness curves. Under this assumption, the mechanism becomes kinematically equivalent to a twisted string actuator (TSA) \cite{popov2012study}, as illustrated in Fig. \ref{fig:fig6}.

The axial contraction $X$ as a function of rotation angle $\theta$ is given by
\begin{align}
X = L - \sqrt{L^2-(R\theta)^2}.
\label{eqn:eq12}
\end{align}
Differentiation yields the velocity relationship
\begin{align}
\dot{X} = \frac{R^2\theta}{\sqrt{L^2-(R\theta)^2}}\dot{\theta}
\label{eqn:eqn13}
\end{align}
and the Jacobian of the TUM,
\begin{align}
J_{TUM} = \frac{\dot{X}}{\dot{\theta}}
= \frac{R^2\theta}{\sqrt{L^2-(R\theta)^2}}.
\label{eqn:eqn14}
\end{align}
Finally, the contraction force $F_c$, elastic force $F_s$, and motor input torque $\tau_{in}$ are related through virtual work and force equilibrium as
\begin{align}
F_c = J_{TUM}^{-1}\tau_{in} - F_s.
\label{eqn:eqn15}
\end{align}

This kinematic and elastic formulation provides a unified framework for predicting motion transmission, force generation, and torque requirements of the TUM within the SPINE gripper system.

\begin{figure*}[!t]
\centering
\includegraphics[width=17cm]{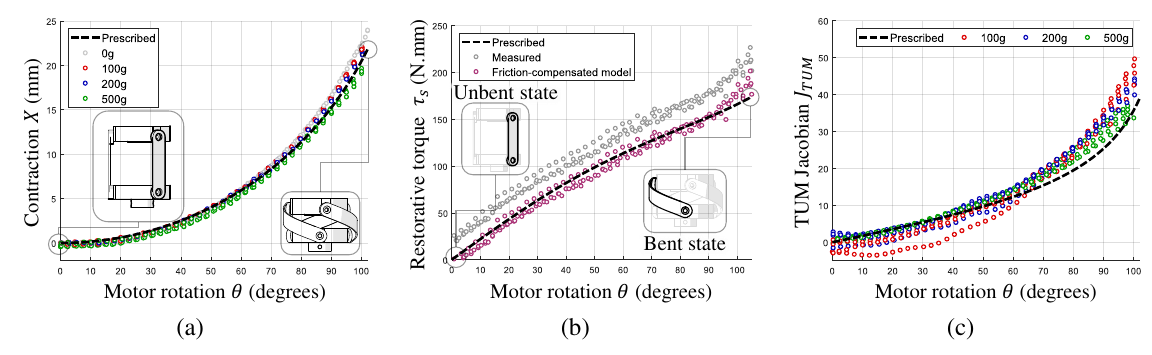}
\caption{Experimental validation of the analytical model of the Twisted Underactuated Mechanism (TUM). (a) Motor rotation–contraction kinematics under different loads, showing close agreement with analytical predictions and reducing deviation at higher loads. (b) Restorative torque of compliant strips aligns with the model after compensating for frictional losses. (c) Experimental Jacobian values compared with theory, with higher loads improving correspondence by suppressing geometric instability.}
\label{fig:fig8}
\end{figure*}

\section{Experimental Evaluation of the Twisted Underactuated Mechanism} \label{sec:section4}

The analytical models of the Twisted Underactuated Mechanism (TUM) predict its rotation--contraction kinematics, strip elasticity, and Jacobian-based transmission behavior. We experimentally validate that the two-dimensional formulations from Section \ref{sec:section3-3} capture the physical mechanism under realistic torque inputs and external loads, thereby establishing the empirical basis for subsequent SPINE gripper characterization.

Fig. \ref{fig:fig7} shows the test setup. A Dynamixel motor (XM590-W290-R, Robotis) drove the TUM; a torque sensor (TRS605, FUTEK) mounted on the output shaft measured input torque and rotation angle, and was coupled directly to the TUM bottom plate. The top plate was constrained to linear motion using a linear guide and connected to a linear encoder (PZ34-A-150, GEFRAN) to measure contraction. A constant downward load was applied via a suspended weight and string. All trials were conducted at 60 rpm with four repetitions per condition.

\subsection{Evaluation of the Twisted Underactuated Mechanism Kinematics} \label{sec:section3-7}

We first verify the TUM kinematics to confirm that the fabricated mechanism follows the predicted mapping from motor rotation angle to axial contraction, which directly governs the grasp-closing motion in the SPINE gripper. Rotation and contraction were measured under four loads (0 g, 100 g, 200 g, and 500 g). In Fig. \ref{fig:fig8}(a), the analytical displacement curve is shown by the black dashed line and experimental data by colored markers.

Across conditions, deviation from theory increases with rotation angle, but decreases monotonically with increasing load. Under 500 g, contraction closely matches the analytical curve over the full range. The larger errors at low loads are attributed to slight warping of compliant strips, which is suppressed under higher tensile loading that improves geometric stability. Thus, under operationally relevant loads, the TUM reproduces the intended kinematics, supporting the validity of the analytical model for predicting SPINE gripper motion.

\subsection{Evaluation of the Compliant Strip Elastic Force} \label{sec:section3-8}

Given that the mode transition and force regulation of the SPINE gripper depend on strip elasticity, we next validate the analytical prediction of restorative torque. Restorative torque was measured as a function of motor rotation under no external load, providing a direct comparison to the model-derived torque profile.

The compliant strips were fabricated using an FDM 3-D printer (F170, Stratasys Ltd.) with ABS (Young's modulus 1200 MPa; second moment of area 1.71 mm$^3$). In Fig. \ref{fig:fig8}(b), measured torque (gray points) follows the theoretical trend (black dashed line) but exhibits an approximately constant offset of 25 N$\cdot$mm, attributable to friction at mechanism and test interfaces. After compensating for this loss, the corrected torque (purple) aligns closely with the analytical prediction, reaching approximately 175 N$\cdot$mm at 105$^\circ$. This agreement confirms that the fabricated strips provide the intended elastic response assumed in the SPINE gripper design.

\subsection{Evaluation of the Twisted Underactuated Mechanism Jacobian} \label{sec:section3-8}

Finally, we validate the Jacobian model, which characterizes the transmission relationship between input torque $\tau_{in}$ and output force and thus determines achievable grasping force and manipulation performance. We measured motor rotation angle and input torque under constant loads of 100 g, 200 g, and 500 g, and computed experimental Jacobian values for comparison with theory (Fig. \ref{fig:fig8}(c)).

Using the force-equilibrium form of (\ref{eqn:eqn15}), $F_s$ can be expressed via the elastic torque $\tau_s$ and the inverse Jacobian. Rearranging yields
\begin{align}
J_{TUM} = \frac{\tau_{in}-\tau_s}{F_s}
\label{eqn:eqn140}
\end{align}
where $\tau_{in}$ and $\tau_s$ are obtained from measured torque and the calibrated elastic response, and $F_s$ is set by the applied load. The experimental Jacobian follows the analytical curve (black dashed line), with discrepancy increasing at larger rotation angles but decreasing with higher load, consistent with the kinematic results. The improved correspondence at higher loads is again explained by suppression of strip warping and reduced geometric instability. Overall, these results confirm that the analytical Jacobian accurately captures the TUM transmission characteristics under realistic loading, supporting its use for predicting SPINE gripper behavior.

\section{Design of the SPINE Gripper} \label{sec:section5}

\subsection{Kinematic Modeling of the SPINE Gripper} \label{sec:section4-1}

\begin{figure}[!t]
\centering
\includegraphics[width=6cm]{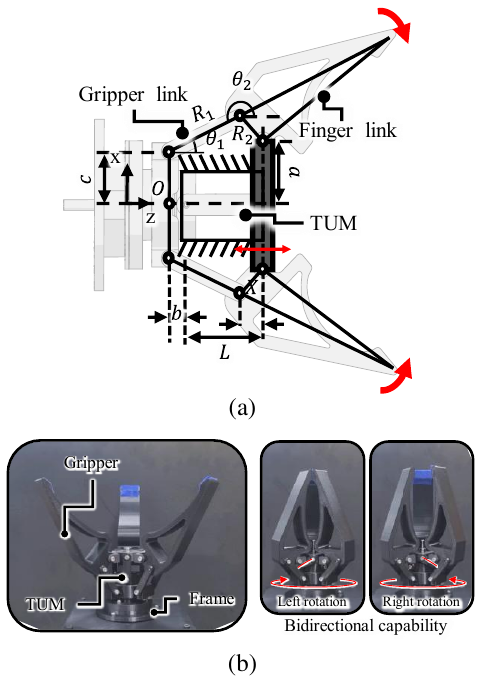}
\caption{Kinematic schematic and fabricated prototype of the SPINE gripper. (a) Schematic for kinematic modeling. (b) Fabricated prototype.}
\label{fig:fig90}
\end{figure}  

Building upon the validated TUM model, this subsection derives the kinematic and force transmission relationships that map the contraction of the Twisted Underactuated Mechanism (TUM) to finger rotation and grasping torque in the SPINE gripper. These relationships directly determine the achievable grasp force under a given input torque.

The gripper employs a slider–crank mechanism driven by the linear contraction of the TUM, as illustrated in Fig. \ref{fig:fig90}(a). In the kinematic model, the TUM is represented as a slider with contraction length $X$. The coordinate system is defined with respect to the origin $O$, where $\theta_1$ denotes the angle between the $z$-axis and the gripper link, and $\theta_2$ denotes the angle between the $z$-axis and the finger link. Geometric parameters $a$, $b$, and $c$ define the horizontal and vertical offsets of the linkage, and $L$ is the initial TUM length.

Applying the kinematic loop equation yields
\begin{align}
\begin{split}
    R_1 (\cos \theta_1 + i \sin \theta_1) 
    + R_2 (\cos \theta_2 + i \sin \theta_2) \\
    - ai - b - L + X + ci = 0
\end{split}
\label{eqn:eqn16}
\end{align}
which can be separated into real and imaginary components:
\begin{align}
R_1 \cos \theta_1 + R_2 \cos \theta_2 - b - L + X &= 0
\label{eqn:eqn17}
\end{align}
\begin{align}
R_1 \sin \theta_1 + R_2 \sin \theta_2 - a + c &= 0.
\label{eqn:eqn18}
\end{align}
Solving (\ref{eqn:eqn17})–(\ref{eqn:eqn18}) yields $\theta_1(X)$ and $\theta_2(X)$, establishing the mapping between TUM contraction and finger rotation.

The finger Jacobian $J_g$, relating contraction velocity to finger angular velocity, is defined as
\begin{align}
J_{g} = \frac{d\theta_2(X)}{dX} = \dot{\theta}_2(X) = \dot{\theta}_2(\theta).
    \label{eqn:eqn19}
    \end{align}
Combining this Jacobian with the TUM force balance from (\ref{eqn:eqn15}), the torque applied at the finger link is expressed as
\begin{align}
\tau_g = J_g \cdot F_c
= J_g \cdot \left( J_{TUM}^{-1} \tau_{input} - F_s \right)
\label{eqn:eqn21}
\end{align}
which directly links motor input torque to grasping torque at the fingertip. This formulation provides a quantitative basis for predicting grasp force and for tuning the gripper geometry and friction threshold.

\subsection{Prototype Fabrication of the SPINE Gripper} \label{sec:section4-2}

\begin{figure*}[!t]
\centering
\includegraphics[width=16.5cm]{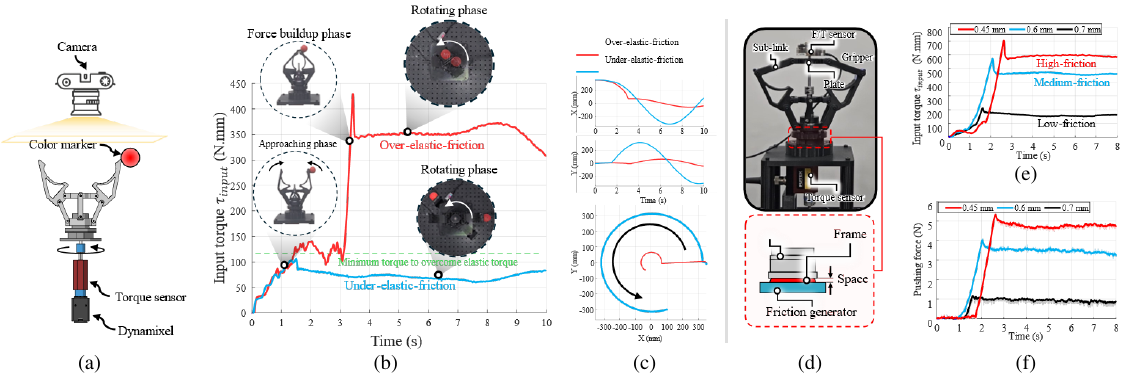}
\caption{Experimental validation of grasp success and friction-based grasping force regulation in the SPINE gripper. (a) Experimental setup for grasp success evaluation. (b) Time–torque profiles under high and low friction, showing approaching, force buildup, and rotation phases. (c) Tracked fingertip trajectories indicating successful grasp (high friction) and incomplete grasp (low friction). (d) Experimental setup for grasping force regulation via friction generator compression. (e) Input torque profiles for three friction conditions set by spacer thickness. (f) Corresponding pushing force profiles demonstrating mechanically tunable and stable grasping force.}
\label{fig:fig13}
\end{figure*}

\begin{table}[t] \label{tab:table1}
\centering
\caption{Specifications of SPINE gripper}
\begin{tabular}{l l}
\hline
\multicolumn{2}{c}{\textbf{Twisted underactuated mechanism (TUM)}} \\
\hline
R (Radius) & 20 mm \\
L (Strip length)  & 40 mm \\
Maximum contraction length & 40 mm \\
Maximum rotation angle & $105^{\circ}$ \\
Material & Acrylonitrile Butadiene Styrene (ABS) \\
\hline
\multicolumn{2}{c}{\textbf{Gripper}} \\
\hline
a & 30 mm \\
b & 10 mm \\
c & 25 mm \\
$R_1$ (Gripper link) & 40 mm \\
$R_2$ (Finger link) & 25 mm \\
Motion range of finger link & $0^{\circ}$ to $120^{\circ}$ \\
Material & Acrylonitrile Butadiene Styrene (ABS) \\
\hline
\multicolumn{2}{c}{\textbf{Friction generator (O-ring)}} \\
\hline
Outer diameter & 35 mm \\
inner diameter & 24.7 mm \\
Thickness & 10.8 mm \\
material & Nitrile rubber (NBR) \\
\hline
\end{tabular}
\end{table}

Based on the derived kinematic and force models, a functional prototype of the SPINE gripper was fabricated to validate the design. The gripper consists of four main components: the TUM, the gripper linkage, a fixed frame, and a friction generator. The system was designed to grasp objects of varying shapes with characteristic dimensions up to 10 cm.

To satisfy the non-interference condition in (\ref{eqn:eqn3}) while ensuring sufficient contraction, the TUM was designed with a strip length of 40 mm and a plate radius of 20 mm. The resulting kinematics allow a finger rotation range of approximately $0^{\circ}$$\sim$$120^{\circ}$, suitable for diverse object sizes. The gripper link lengths were set to $R_1 = 40$ mm and $R_2 = 16.25$ mm, and the linkage offsets were chosen as $a = 30$ mm, $b = 10$ mm, and $c = 25$ mm to accommodate mechanical constraints. All design parameters are summarized in Table~\uppercase\expandafter{\romannumeral1}.

All structural components were manufactured using an FDM-type 3-D printer (F170, Stratasys Ltd.) with ABS material. The friction generator, which determines the maximum grasping force, was implemented using a nitrile rubber (NBR) O-ring with an outer diameter of 35 mm, an inner diameter of 24.7 mm, and a thickness of 10.8 mm. The O-ring is inserted between the gripper and frame, creating a frictional interface whose compression distance directly sets the grasping force limit. This simple mechanical element enables tunable and repeatable force regulation without sensors or active control, as validated experimentally in Fig. \ref{fig:fig90}(b).

\section{Experimental Evaluation of the SPINE Gripper prototype} \label{sec:section6}

\subsection{Grasp Success Condition and Results} \label{sec:section5-1}

This experiment evaluates whether the SPINE gripper can reliably achieve a full grasp prior to in-hand rotation, focusing on the interaction between the elastic force generated by the TUM and the frictional resistance provided by the friction generator. Successful grasping requires that the maximum static friction exceed the restorative elastic torque of the TUM; otherwise, slip occurs before sufficient contraction is transmitted to the gripper.

Fig. \ref{fig:fig13}(a) illustrates the experimental setup. A Dynamixel motor (XM590-W290-R, Robotis) applied input torque, which was measured in real time using a torque sensor (TRS605, FUTEK). Grasp success was defined as the gripper reaching a fully closed configuration. Fingertip displacement was tracked using a color marker and a top-view camera operating at 60 FPS, with trajectories extracted via OpenCV-based marker tracking.

The TUM required approximately 115 N$\cdot$mm of input torque to overcome its internal elastic force in the absence of external load. Two friction conditions were tested: a high-friction condition, in which the maximum static friction exceeded the elastic threshold, and a low-friction condition, in which it did not.

Measured torque profiles are shown in Fig. \ref{fig:fig13}(b). Under the high-friction condition (red), the input torque exhibited three distinct phases: (i) an approaching phase with gradual torque increase and contraction onset, (ii) a force-buildup phase following fingertip contact, reaching a static friction peak of approximately 440 N$\cdot$mm, and (iii) a rotating phase after slip, with torque stabilizing near the kinetic friction level of approximately 340 N$\cdot$mm. In contrast, under the low-friction condition (blue), the static friction limit (approximately 100 N$\cdot$mm) was lower than the elastic threshold. As a result, the gripper transitioned directly to rotation during the approaching phase, stabilizing near 90 N$\cdot$mm without achieving full closure.

Fingertip trajectories in Fig. \ref{fig:fig13}(c) further confirm this behavior. Under high friction, the fingertip moved from an initial position of 320 mm to approximately 50 mm along the X-axis, indicating full closure before rotation, with minimal lateral displacement due to the small rotation radius. Under low friction, rotation initiated prematurely at approximately 300 mm, resulting in a larger rotation radius and significantly increased X–Y displacement. These results demonstrate that a sufficient friction threshold is essential for achieving stable grasping prior to rotation.

\begin{figure*}[!t]  
\centering
\includegraphics[width=16cm]{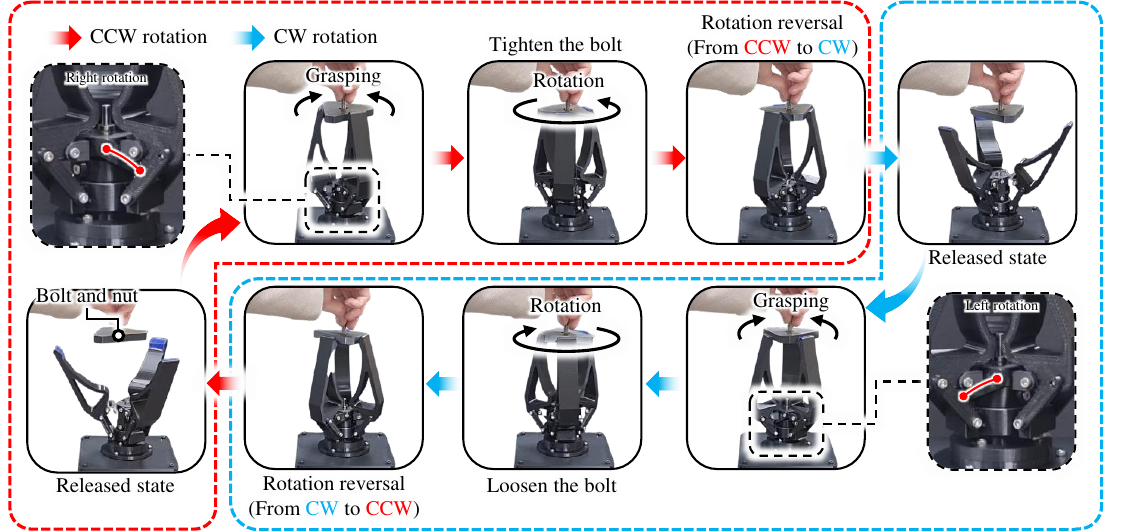}

\caption{Demonstration of the SPINE gripper's bidirectional capability. The gripper performs grasping and in-hand rotation in both CCW and CW directions through mechanically encoded power transmission logic, without sensors or additional actuators. Alternately loosening and tightening a bolt demonstrates stable grasp–release transitions and continuous rotation, confirming bidirectional manipulation enabled solely by structural design.}
\label{fig:fig1501}
\end{figure*}

\subsection{Friction and Grasping Force Evaluation} \label{sec:section5-2}

Building on the grasp success analysis, this subsection evaluates how friction generator compression regulates grasping force and its stability during rotation. Spacer thicknesses of 0.45 mm, 0.6 mm, and 0.7 mm were used to systematically vary the friction level.

The experimental setup is shown in Fig. \ref{fig:fig13}(d). Input torque and rotation angle were measured using a torque sensor (TRS605, FUTEK). A force/torque sensor (ATI Mini45) was mounted coaxially with the gripper via an extended shaft and a single-DOF sub-link. During grasping, a plate attached to the sub-link translated along the shaft and pushed against the sensor. Due to the slider–crank kinematics, the measured pushing force has a one-to-one relationship with the fingertip grasping force.

Time histories of input torque and grasping force are shown in Fig. \ref{fig:fig13}(e) and (f), with raw data filtered using a moving average window of 10 samples. Under the high-friction condition (0.45 mm spacer), the maximum static and kinetic friction torques were approximately 700 N$\cdot$mm and 560 N$\cdot$mm, respectively, corresponding to pushing forces of 5.3 N (static) and 4.6 N (rotation). With a 0.6 mm spacer, static and kinetic friction torques decreased to approximately 568 N$\cdot$mm and 462 N$\cdot$mm, yielding pushing forces of 4.0 N and 3.5 N. Under the low-friction condition (0.7 mm spacer), the friction torques dropped to approximately 210 N$\cdot$mm (static) and 170 N$\cdot$mm (kinetic), with corresponding pushing forces of 1.17 N and 1.05 N.

These results demonstrate that the SPINE gripper’s maximum grasping force can be mechanically tuned through friction generator compression, without sensors or active force control. Across all conditions, once the friction threshold was reached, both torque and grasping force remained stable during continuous in-hand rotation. This confirms that the gripper maintains reliable and predictable grasping performance while passively transitioning between grasping and rotation.

\section{Demonstration} \label{sec:section7}

\begin{figure*}[!t]  
\centering
\includegraphics[width=17cm]{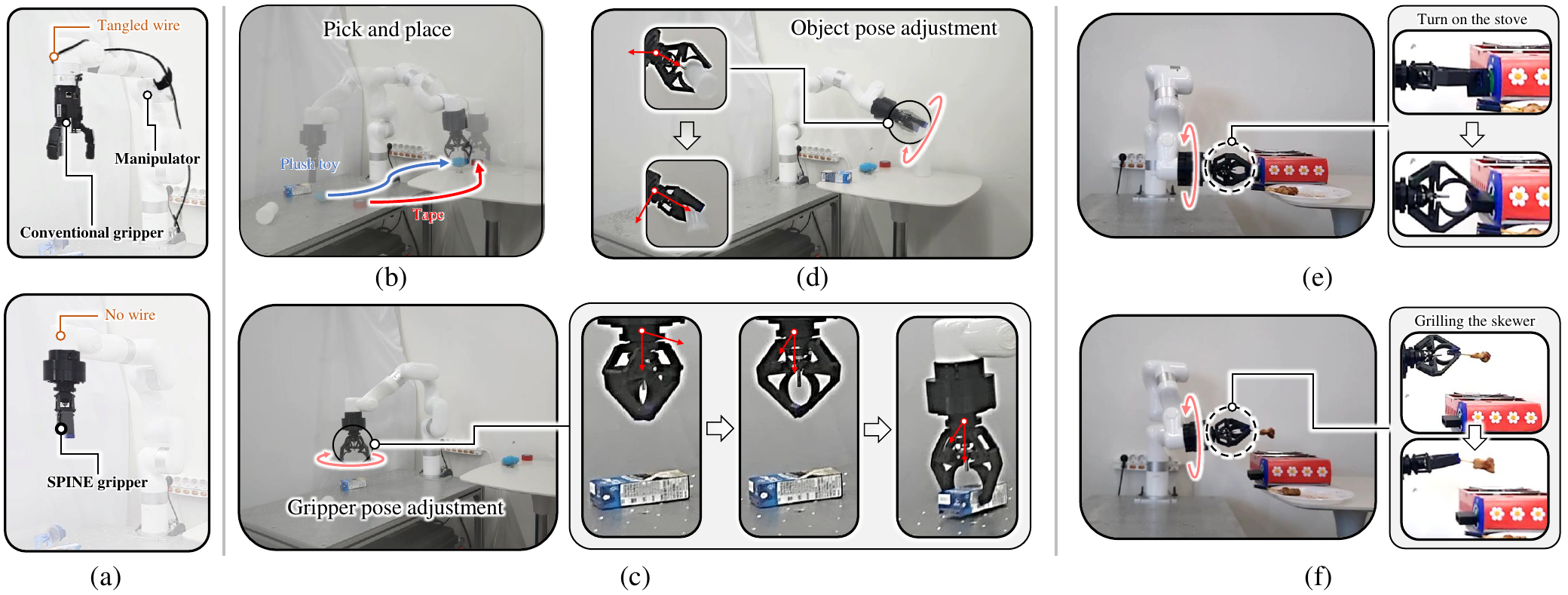}

\caption{System-level evaluation of the SPINE gripper integrated with a robotic manipulator, driven solely by wrist torque. (a) Comparison between a conventional gripper with electrical wiring constraints (twisting, interference, rotation limits) and the SPINE gripper operating without motors, sensors, or wiring. (b) Pick-and-place demonstrations with rigid and deformable objects using only wrist torque. (c) Pose adjustment through wrist rotation to acquire grasp when initial object orientation is unfavorable. (d) Object reorientation via in-hand rotation for alignment during insertion into a narrow opening. (e) Kitchen demonstration: turning a stove knob and (f) rotating a skewer over a grill, confirming real-world manipulation without motors or sensing hardware.}
\label{fig:fig1500}
\end{figure*}

\subsection{Bidirectional capability of the SPINE Gripper} \label{sec:section6-1}

Bidirectional in-hand rotation is essential for manipulation tasks that require both stable grasping and controlled object rotation. While many underactuated grippers conform to object geometry, few can achieve continuous rotation in both directions without additional actuators or control complexity. Demonstrating this capability is therefore critical to validating the core premise of the SPINE gripper: that grasping and rotation are passively coordinated through mechanically encoded logic.

To evaluate this function, a bolt loosening–tightening task was performed, which explicitly exposes the transition between grasping and rotation (Fig. \ref{fig:fig1501}). Starting from an open state, a counterclockwise (CCW) input causes the gripper to grasp the nut; once the transmitted torque exceeds the static friction threshold, the mechanism passively transitions to rotation and loosens the bolt (red dashed box). Reversing the input direction releases the nut and resets the mechanism. Continued clockwise (CW) rotation then induces a second grasp followed by tightening of the bolt (blue dashed box).

All transitions between grasping, release, and rotation occurred solely through the mechanical thresholds embedded in the structure, without sensing, closed-loop control, or additional actuation. The ability to repeatedly perform loosening and tightening with a single actuator confirms that the bidirectional behavior arises from the mechanism itself rather than from external control strategies. This result demonstrates a level of functionality that is difficult to achieve with conventional underactuated grippers and validates the effectiveness of mechanically encoded power transmission logic.

\subsection{Implementation the SPINE Gripper on a Robotic Manipulator}\label{sec:section6-2}

While isolated mechanism tests confirm functional principles, practical deployment requires evaluation at the system level. In particular, integration with a robotic manipulator must address installation complexity, interaction with joint motion, and wiring-related constraints. This subsection therefore evaluates the SPINE gripper when mounted on a conventional manipulator and driven exclusively by wrist torque.

As shown in Fig. \ref{fig:fig1500}(a), conventional grippers typically rely on motors and sensors integrated within the gripper, which require electrical wiring to be routed through multiple manipulator joints. Such wiring is prone to fatigue, interference, and twisting during wrist rotation, which limits continuous rotation and increases maintenance demands. In contrast, the SPINE gripper contains no motors, sensors, or wiring at the end effector (Fig. \ref{fig:fig1500}(a)). When integrated with a manipulator, the SPINE gripper utilizes the wrist's rotational motion as the input for grasping (Fig. \ref{fig:fig17}). Once the grasping force reaches a certain threshold, the system transitions to enable the manipulator wrist's original functionality (in-hand rotation). This design enables both grasping and wrist rotation without requiring any sensors or motors in the gripper, thereby eliminating common failure modes associated with cables and electronic components. This structural simplicity enables a degree of rotational freedom not readily achievable in motor-driven grippers.

The first system-level demonstration evaluated basic pick-and-place operations (Fig. \ref{fig:fig1500}(b)). A rigid tape roll and a deformable plush doll were selected to represent different contact conditions. In both cases, the 6-DOF serial manipulator (xArm 6, UFactory)  applied wrist torque to grasp, confirming that the SPINE gripper can adapt to varying object properties while maintaining sufficient stability for manipulation.

The second demonstration examined pose adjustment during manipulation, which is essential when object orientation is unfavorable. As shown in Fig. \ref{fig:fig1500}(c), the manipulator rotated its wrist to reposition the gripper before grasping a milk carton placed at an awkward angle. In a related task (Fig. \ref{fig:fig1500}(d)), the gripper rotated a grasped object during placement to align it with a narrow opening. These scenarios demonstrate that wrist-driven in-hand rotation enables orientation adjustment without additional actuators.

To assess performance in realistic tasks, kitchen demonstrations were conducted. In Fig. \ref{fig:fig1500}(e), the gripper grasped and rotated a stove knob to ignite a burner, requiring secure grasping followed by continuous rotation. In addition, the gripper grasped a skewer and rotated it over a grill while maintaining stable contact (Fig. \ref{fig:fig1500}(f)). Both tasks were performed using only wrist torque, illustrating that the passive mechanism can execute coordinated grasp-and-rotate actions typical of daily manipulation.

Finally, a smart farm demonstration evaluated fruit harvesting (Fig. \ref{fig:fig17}). The manipulator positioned the gripper around a fruit, grasped it using wrist torque, and then rotated it to twist the fruit from the stem. The gripper maintained stable contact throughout the motion, allowing the fruit to detach naturally. This sequence, consisting of approach, grasp, rotate, and detach, was completed without sensors, motors, or power units at the tool tip.

Across all demonstrations, the SPINE gripper consistently performed tasks that typically require electrically actuated end effectors, despite having no active components at the tool tip. The gripper maintained stable grasping, enabled continuous in-hand rotation, and operated reliably across diverse environments. These results confirm that mechanically encoded power transmission logic alone support a wide range of manipulation tasks while eliminating common failure modes related to wiring, sensing, and actuation, highlighting the SPINE gripper’s suitability for robust real-world deployment.

\section{Conclusion}
\label{sec:section8}
This study explored how mechanically encoded power transmission logic can be leveraged to achieve multifunctional manipulation without increasing actuation or control complexity. By embedding decision logic directly into the mechanical structure, the proposed SPINE gripper demonstrates that stable grasping and continuous bidirectional in-hand rotation are unified within a single passive actuation pathway, despite being inherently non-coplanar and traditionally conflicting manipulation functions.

A central outcome of this work is the realization that direction-invariant force generation, achieved through the twisted underactuated mechanism, fundamentally changes the design space of passive grippers. Rather than treating mode switching as a control or sensing problem, this approach reframes it as a mechanical transmission problem, where functional transitions emerge naturally from the interaction between torque, compliance, and friction. This perspective distinguishes the proposed gripper from prior single-actuator designs that remain constrained by unidirectional operation or require explicit reset mechanisms.

Beyond the specific gripper implementation, the results suggest a broader implication for robotic mechanism design: complex manipulation behaviors need not be orchestrated exclusively through software or additional hardware, but can instead be encoded into the physical structure itself. Such an approach offers inherent robustness, simplicity, and scalability, particularly in contact-rich or resource-constrained environments where sensing and control are costly or unreliable.

In this sense, the SPINE gripper serves not only as a functional end-effector, but also as an example of how mechanical design can actively participate in decision-making, opening new avenues for passive and underactuated robotic systems that prioritize reliability and functional elegance over complexity.

\bibliographystyle{IEEEtran}
\bibliography{ref.bib}

\vfill

\begin{IEEEbiography}[{\includegraphics[width=1in,height=1.25in,clip,keepaspectratio]{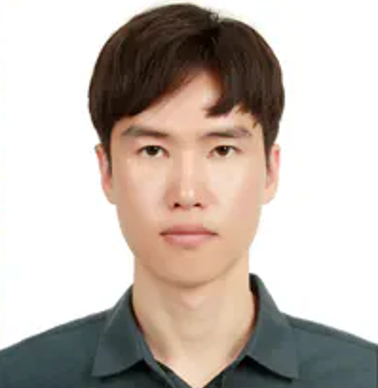}}]{JaeHyung Jang}
received the B.S. degree from Mechanical Engineering in Yeungnam University, Gyeongsangbuk-do, South Korea in 2016, 
and M.S degree from University of Science and Technology, Daejeon, South Korea in 2018, 
and Ph.D. degrees from the Korea Advanced Institute of Science and Technology, Daejeon, South Korea, in 2025. 

He is currently a Postdoc with the Department of Civil and Environmental Engineering, Korea Advanced Institute of Science and Technology. His research interests include Underactuated mechanism design, passive systems, actuator design, and Novel hardware mechanism design.
\end{IEEEbiography}

\begin{IEEEbiography}[{\includegraphics[width=1in,height=1.25in,clip,keepaspectratio]{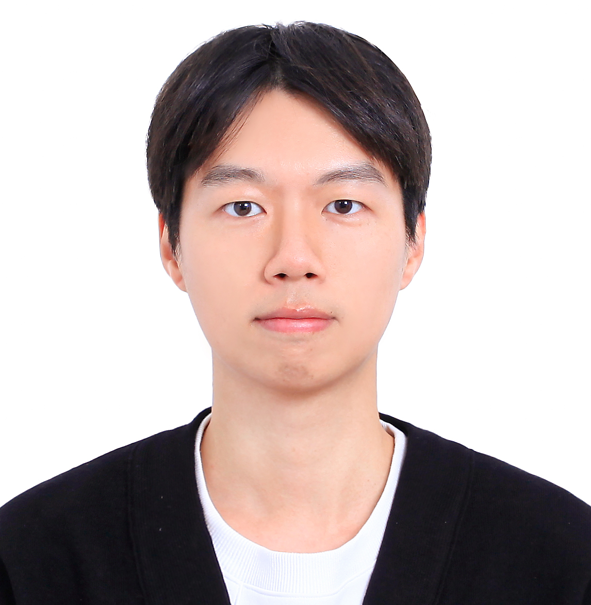}}]{JunHyeong Park}
JunHyeong Park received the B.S. degrees in Civil, Environmental and Architectural Engineering and Mechanical Engineering from Korea University, Seoul, South Korea, in 2026.

He is currently working toward the M.S degree in Robotics with the Korea Advanced Institute of Science and Technology (KAIST), Daejeon, South Korea. His current research interests include actuator design and novel hardware mechanism design.   
\end{IEEEbiography}

\begin{IEEEbiography}[{\includegraphics[width=1in,height=1.25in,clip,keepaspectratio]{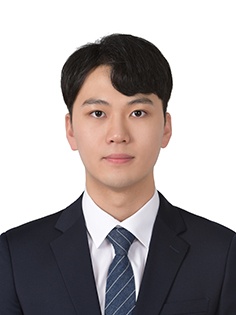}}]{Joong-Ku Lee}
(Student Member, IEEE) received the B.S. degree in Mechanical Engineering from Sungkyunkwan University, Suwon, South Korea, in 2020, and the M.S. degree in robotics from the Korea Advanced Institute of Science and Technology (KAIST), Daejeon, South Korea, in 2022. 

He is currently working toward the Ph.D. degree in Civil and Environmental Engineering at KAIST. His research interests include telerobotics, haptics, and artificial intelligence.
\end{IEEEbiography}

\begin{IEEEbiography}[{\includegraphics[width=1in,height=1.25in,clip,keepaspectratio]{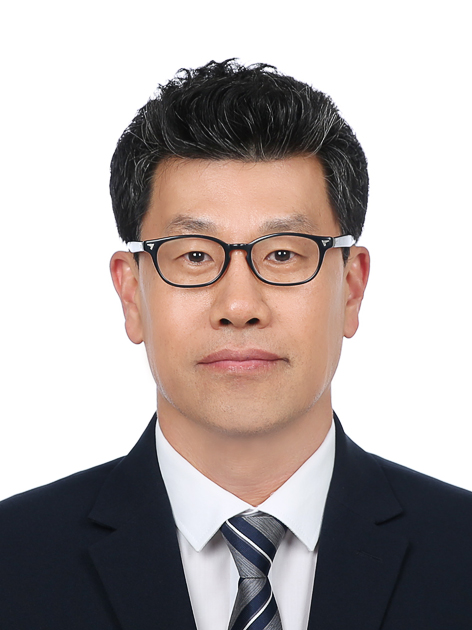}}]{Jee-Hwan Ryu}
(Senior Member, IEEE) received the B.S. degree from Inha University, Incheon, South Korea, in 1995, and the M.S.
and Ph.D. degrees from the Korea Advanced Institute of Science and Technology, Daejeon, South Korea, in 1997 and 2002, respectively, all in mechanical engineering.

He is currently a Professor with the Department of Civil and Environmental Engineering, Korea Advanced Institute of Science and Technology. His research interests include haptics, telerobotics, teleoperation, exoskeletons, and autonomous vehicles.
\end{IEEEbiography}

\end{document}